\documentclass[graybox]{svmult}

\usepackage{mathptmx}       
\usepackage{helvet}         
\usepackage{courier}        
\usepackage{type1cm}        

\usepackage{makeidx}         
\usepackage{graphicx}        
\usepackage{multicol}        
\usepackage[bottom]{footmisc}

\usepackage[T1]{fontenc}
\usepackage[numbers,sort&compress]{natbib}
\usepackage{verbatim}
\usepackage{url}
\usepackage{amsmath}
\usepackage{amsfonts}
\usepackage{enumitem}
\usepackage{amssymb}
\makeindex             

\begin{document}

\title*{Dominance-based Rough Set Approach, basic ideas and main trends}
\titlerunning{DRSA, basic ideas and main trends}
\author{Jerzy B{\l}aszczy\'nski, Salvatore Greco, Benedetto Matarazzo, and Marcin Szel\k{a}g}
\institute{Jerzy B{\l}aszczy\'nski \at Poznan Supercomputing and Networking Center, 61-139~Pozna\'n,~Poland; \email{jurekb@man.poznan.pl}
\and Salvatore Greco \at Department of Economics and Business, University of Catania, 95129 Catania,~Italy\\
  and\\
Centre for Operational Research \& Logistics, Portsmouth Business School, PO1 3DE Portsmouth, UK; \email{salgreco@unict.it}
\and Benedetto Matarazzo \at Department of Economics and Business, University of Catania, 95129 Catania,~Italy; \email{matarazz@unict.it}
\and Marcin Szel\k{a}g \at Institute of Computing Science, Pozna\'n University of Technology, 60-965~Pozna\'n,~Poland; \email{marcin.szelag@cs.put.poznan.pl}}


\maketitle

\abstract*{Dominance-based Rough Approach (DRSA) has been proposed as a machine learning and knowledge discovery methodology to handle Multiple Criteria Decision Aiding (MCDA). Due to its capacity of asking the decision maker (DM) for simple preference information and supplying easily understandable and explainable recommendations, DRSA gained much interest during the years and it is now one of the most appreciated MCDA approaches. In fact, it has been applied also beyond MCDA domain, as a general knowledge discovery and data mining methodology for the analysis of monotonic (and also non-monotonic) data. In this contribution, we recall the basic principles and the main concepts of DRSA, with a general overview of its developments and software. We present also a historical reconstruction of the genesis of the methodology, with a specific focus on the contribution of Roman S{\l}owi\'nski.}

\abstract{Dominance-based Rough Approach (DRSA) has been proposed as a machine learning and knowledge discovery methodology to handle Multiple Criteria Decision Aiding (MCDA). Due to its capacity of asking the decision maker (DM) for simple preference information and supplying easily understandable and explainable recommendations, DRSA gained much interest during the years and it is now one of the most appreciated MCDA approaches. In fact, it has been applied also beyond MCDA domain, as a general knowledge discovery and data mining methodology for the analysis of monotonic (and also non-monotonic) data. In this contribution, we recall the basic principles and the main concepts of DRSA, with a general overview of its developments and software. We present also a historical reconstruction of the genesis of the methodology, with a specific focus on the contribution of Roman S{\l}owi\'nski.}

\section{Introduction}

Among the many merits of Roman S{\l}owi\'nski in his so long and so rich scientific carrier, we have to consider his pioneering approach to the use of artificial intelligence methodologies to decision support, and, in particular, to Multiple Criteria Decision Aiding (MCDA) (for an updated state of the art see \cite{GrecoEhrgottFigueira2016}). In this perspective, the proposal and the development of the Dominance-based Rough Set Approach (DRSA) is a cornerstone in the domain. The DRSA basic idea  of a decision support procedure based on a decision model expressed in natural language and obtained from simple preference information in terms of exemplary decisions has attracted the interest of experts and it is now considered one of the three main approaches to MCDA, together with the classical Multiple Attribute Utility Theory (MAUT) \cite{KeeneyRaiffa1976} and the outranking approach \cite{Roy1991}. In fact, DRSA is not a mere application to MCDA of concepts and tools already proposed and developed in the domain of artificial intelligence, knowledge discovery, data mining and machine learning. Indeed, consideration of preference orders typical for MCDA problems required a reformulation of many important concepts and methodologies, so that DRSA became a methodology viable and interesting \textit{per se} also in these domains. Consequently, after more or less 25 years from the proposal of DRSA, we try to present a first assessment taking into consideration the basic ideas and the main developments.

This paper is organized as follows. Next section presents some historical notes. Section 3 recalls the basic concepts of DRSA, while section 4 describes the main developments. Section 5 presents some available software. Section 6 collects conclusions.
      
\section{Some historical notes on the Dominance-based Rough Set Approach}
In the beginning of the 80’s of the last century, Roman S{\l}owi\'nski entered in contact with two very relevant figures of researchers in two quite different domains: Bernard Roy and Zdzis{\l}aw Pawlak. Bernard Roy, one of the pioneers of the Operation Research in Europe, was the founder of the European School of Multiple Criteria Decision Aiding (MCDA). Zdzis{\l}aw Pawlak, one of the founders of the computer science and artificial intelligence, proposed the Rough Set Theory as a mathematical tool for data analysis and knowledge discovery. Roman S{\l}owi\'nski, that had already given fundamental contributions in scheduling theory, was enthusiastically interested in both MCDA and rough set theory as witnessed, among the others, by a translation in Polish \cite{roy1990wielokryterialne} of the book in which Bernard Roy systematically presented the basis of MCDA \cite{roy1985methodologie} and by the organization of the first rough set international conference held in 1992 in Pozna\'n (the proceedings of the conference are collected in \cite{slowinski1992intelligent}). Since his first contributions in the domain, Roman S{\l}owi\'nski was interested in the use of rough set theory for decision support. In particular, he proposed the first application in real world problems related to application of rough set theory to medical diagnosis \cite{slowinski1988rough}. Moreover, he realized that very interesting developments could be obtained in rough set theory by applying concepts proposed in MCDA, as the construction of similarity relations using indifference thresholds \cite{slowinski1992generalization}, inspired by analogous concepts in outranking methods \cite{roy1984relational}. Pursuing this research line, very soon Roman S{\l}owi\'nski matured the conviction that beyond the simple application in rough set theory of some specific concepts originating in MCDA, there was the space for a whole extension of rough set theory that could become a relevant model for  MCDA \cite{ pawlak1994decision,pawlak1994rough}. 
In May 1994 Roman S{\l}owi\'nski went to Catania to give some seminars on MCDA and rough set theory. On that occasion, discussing with  Benedetto Matarazzo and Salvatore Greco, a very interesting idea came out: what could be obtained approximating sets of pairs of objects (binary relations) rather than ``standard'' sets of single objects? The information to be processed with rough set theory  had to be the preference on single criteria and the overall preference, in order to express the overall preference in terms of preference on the single criteria. This idea was called rough set analysis of a pairwise comparison table (PCT). In October of the same year Roman S{\l}owi\'nski, Salvatore Greco and Benedetto Matarazzo met again at the 40th meeting of the EURO working group on Multiple Criteria Decision Aiding, held in Paris and Bordeaux. The program of the meeting was split in two parts, with the morning of the 6th of October in Paris and  the following day in Bordeaux, with a transfer by train in the afternoon of the 6th of October. The travel between Paris and Bordeaux was a very good occasion for a long discussion on the intuition that come out in Catania.  The three researchers remained in contact continuing to work on PCT. On the first days of May 1995, Salvatore Greco stayed one week in Pozna\'n and during that week the first paper on PCT was completed \cite{greco1997rough}. In the same year the new rough set model was presented at an international conference, more precisely at the  12th International Conference on Multiple Criteria Decision Making held on June 19-23 in Hagen \cite{fandel1997multiple}. This was the first presentation of PCT at an MCDA conference. The year after, Roman S{\l}owi\'nski, Salvatore Greco and Benedetto Matarazzo participated in the  Fourth International Workshop on Rough Sets, Fuzzy Sets, and Machine Discovery, held on November 6-8, 1996 in Tokyo. This was the first exposition of the PCT to a rough set conference.  On that occasion, Hirotaka Nakayama invited the three researchers to Kobe.  Again during the travels in train between Tokyo and Kobe and back there were very rich and constructive discussions about the new rough set model. Both researchers in MCDA and in rough set theory were interested in the idea. However, very soon it appeared clear that the PCT was not able to represent all the salient aspects of MCDA. Indeed, classical rough set theory is based on indiscernibility relation, so what is taken into account is if the considered objects have the same description or not. In MCDA, and, more in general, in decision support, there is something more to be considered than the equal or different description. For example, in deciding about cars, a maximum speed of 200 km/h is greater and consequently preferred to a maximum speed of 180 km/h. However, under the lens of indiscernibility 200 km/h is only different from 180 km/h. How to take into account that beyond the difference there is the preference of 200 km/h over 180 km/h? For a certain time this was not clear and there was even a point in which Roman S{\l}owi\'nski, Salvatore Greco and Benedetto Matarazzo took into consideration to abandon the research. In the same period, Constantin Zopounidis invited Roman S{\l}owi\'nski to write a contribution for a book that he was editing \cite{zopounidis1998operational}. Roman S{\l}owi\'nski proposed Salvatore Greco and Benedetto Matarazzo to write this contribution by applying their approach to rough set theory to bankruptcy evaluation, using the data that had been analyzed with classical rough set theory in \cite{dimitras1999business}. Working on these data, the idea that would permit to extend the rough set theory to MCDA was finally found. The key concept was the substitution of the indiscernibility relation used in classical rough set theory with the dominance relation on which MCDA is rooted. On the basis of this intuition, definitions and results of classical rough set theory can be reformulated in the MCDA perspective. This idea quickly proved to be very successful in improving the results obtained using classical rough set theory, and the contribution in the book edited by  Constantin Zopounidis \cite{zopounidis1998operational} became the first application of the new extension of rough set theory called Dominance-based Rough Set Approach. The new idea gave new impetus to the research, so that a systematic analysis of the basic concepts of DRSA and of its extension was pursued. At those times, Roman S{\l}owi\'nski received the invitation to write a chapter on fuzzy sets applied to MCDA, for a book proposing a state of the art on MCDA \cite{gal2013multicriteria}. To write that contribution, in the spring 1998 Salvatore Greco stayed two months in Pozna\'n to cooperate with Roman S{\l}owi\'nski, remaining constantly in contact with Benedetto Matarazzo. The final result was a contribution in the book in which rough set theory applied to MCDA, rather than fuzzy set theory, was presented \cite{greco1999use}. The material in that chapter was continuously revised and improved until it became the basic paper in which DRSA is presented \cite{ greco2001rough}. This paper has become one of the most read and cited of the European Journal of Operational Research. The research on DRSA took several directions. Some years after, on December 2006, Roman S{\l}owi\'nski, together with Juergen Branke, Kaylamon Deb and Kaisa Miettinen organized a Dagstuhl seminar with the aim of opening a discussion between researchers interested in interactive multiobjective optimization and  evolutionary multiobjective optimization (the presentations of that seminar are collected in \cite{branke2008multiobjective}). In that seminar, Roman S{\l}owi\'nski and Salvatore Greco presented the result of a new research together with Benedetto Matarazzo. The idea was the application of DRSA to guide the search of the most preferred solution in an interactive multiobjective optimization problem \cite{greco2008dominance}. Again, the proposal was well accepted by the experts in multiobjective optimization, because through the DRSA the preference information asked from the DM is very simple and intuitive (the classification of some solutions as good or not) and the preference model supplied by DRSA in terms of ``if \ldots, then \ldots'' decision rules is very understandable for the DM and easy to be managed in the optimization algorithm. Recently, Roman S{\l}owi\'nski has come back to the multiobjective optimization through DRSA in a new research conducted again with Salvatore Greco and Benedetto Matarazzo with the addition of Salvatore Corrente. The new research \cite{corrente2021explainable} aims at using DRSA in an evolutionary multiobjective optimization algorithm. The use of DRSA is justified taking into account also its good properties from the point of view of decision psychology \cite{lichtenstein2006construction}.

\section{Basic concepts of the Dominance-based Rough Set Approach}\label{sec:basics}

In this section, we want to present the basic ideas of DRSA \cite{greco2001rough} in comparison with the original Indiscernibility-based Rough Set Approach (IRSA) proposed by Zdzis{\l}aw Pawlak \cite{pawlak_1982,pawlak1991rough}. Both IRSA and DRSA consider a universe $U$ being a finite set of {\it objects}, a finite set of {\it attributes} $Q=\{q_1,q_2,\ldots,q_m\}$, each one of them having a value set $V_{q_i}, i=1, \ldots,m$, and an information function $f:U\times Q\to V$, with $V=\bigcup_{q\in Q} V_{q}$, such that $f(x,q)\in V_q$ for each $q\in Q$.

Every set of attributes $P\subseteq Q, P\neq \emptyset$, defines
an {\it indiscernibility relation} on $U$, denoted by $I_P$:
\[
I_P = \big\{(x,y)\in U\times U: f(x,q) = f(y,q),\hbox{for
all} \  q\in P\big\}.
\]

If $(x,y)\in I_P$, denoted also $x I_P y$, we say that the objects $x$ and $y$ are
$P$-indiscernible. The indiscernibility relation $I_P$ is an equivalence relation on $U$ assigning to each object $x \in U$  its equivalence class
\[
I_P(x)=\{y \in U: y I_P x\}.
\]
The family of all the equivalence
classes of relation $I_P$ is denoted by $U|I_P$. The equivalence classes of relation $I_P$
are called the  {\it $P$-elementary sets} or {\it granules
of knowledge} encoded by $P$. 

Using the indiscernibility relation $I_P$, to any set $X\subseteq U$ may be associated
the {\it $P$-lower
approximation} 
\[
\underline P(X) = \big\{x\in U\colon I_P(x)\subseteq
X\big\}
\]
and the {\it $P$-upper approximation}
\[
\overline P(X) = \big\{x\in U\colon I_P(x) \cap X\neq
\emptyset \big\}\,.
\]

Intuitively, an object $x$ belongs to $\underline P(X)$ if it
is  {\it certainly} contained in $X$, in the sense that all the objects that are indiscernible with it also belong to $X$. Instead, an object $x$
belongs to $\overline P(X)$ if it is {\it possibly}
contained in $X$, in the sense that there is at least one object indiscernible with $x$ that belongs to $X$. 

Often the set of attributes  $Q$ is 
divided into the set of {\it condition} attributes 
\mbox{$C\neq\emptyset$} and the set of {\it decision} attributes 
$D\neq\emptyset$, such that $C\cup D=Q$ and $C\cap
D=\emptyset$. The indiscernibility relation $I_S$ with respect to a set of decision attributes $S\subseteq D$, $S\neq \emptyset$,  induces a~partition of $U$, so that the lower and the upper approximation of each equivalence class $I_S(x)$, $x \in U$, with respect to $P \subseteq C$, $P \neq \emptyset$, can be computed, with the aim of discovering dependencies between condition attributes from $P$ and decision attributes from~$S$. Indeed, if for $x,z \in U$, $P\subseteq C$ and $S \subseteq D$, $x \in \underline{P}(I_S(z))$, then $I_P(x) \subseteq I_S(z)$, that is, for all $y \in U$, if $f(y,q)=f(x,q)$ for all $q \in P$, then $f(y,q)=f(z,q)$ for all $q \in S$. This can be interpreted in the sense that the objects from $U$ suggest the following \textit{certain} decision rule:

\begin{center}
$\mathbf{\rho}_c$=``if $f(y,q)=f(x,q)$ for all $q \in P$, then $f(y,q)=f(z,q)$ for all $q \in S$''.
\end{center}

Analogously, if for $x,z \in U$, $P\subseteq C$ and $S \subseteq D$, $x \in \overline{P}(I_S(z))$, then $I_P(x) \cap I_S(z) \neq \emptyset$, that is, there is at least one $y \in U$, such that $f(y,q)=f(x,q)$ for all $q \in P$ and $f(y,q)=f(z,q)$ for all $q \in S$. This can be interpreted in the sense that the objects from $U$ suggest the following \textit{possible} decision rule:

\begin{center}
$\mathbf{\rho}_p$=``if $f(y,q)=f(x,q)$ for all $q \in P$, then it is possible that $f(y,q)=f(z,q)$ for all $q \in S$''.  
\end{center}
The certain and possible decision rules induced from universe $U$ can also be applied to classify objects not in universe $U$, with an easily understandable explanation expressed in natural language. In fact, in presence of a certain decision rule $\mathbf{\rho}_c$ (a~possible decision rule $\mathbf{\rho}_p$),  if there is an object $w \notin U$ such that $f(w,q)=f(x,q)$ for all $q \in P$, the objects from $U$ suggest that one has to (could) expect $f(w,q)=f(z,q)$ for all $q \in S$, with the certain decision rule $\mathbf{\rho}_c$ (the possible decision rule $\mathbf{\rho}_p$) that can be seen as an explanation.

Very often there is a single decision attribute $d$, that is $D=\{d\}$. In this case, the equivalence classes $I_d(x), x \in U$, can be identified with a set of decision classes $\mathbf{Cl}$$=\{Cl_1,\ldots,Cl_n\}$. In this context, the lower and the upper approximation $\underline P(Cl_t)$ and $\overline P(Cl_t)$ of each decision class $Cl_t \in \mathbf{Cl}$ with respect to a set of condition attributes $P \subseteq C, P \neq \emptyset$, can be obtained.

The classical IRSA has been recognized as a mathematical theory useful in tasks considered in knowledge discovery and data mining. It has been widely investigated from the theoretical perspective. It has also been applied in analysis of many real world problems. However, as explained in the historical notes of the previous section, it cannot deal with preferences and, more in general, with data exhibiting monotonic relationships. In order to handle this problem, Roman S{\l}owi\'nski with Salvatore Greco and Benedetto Matarazzo proposed to substitute the indiscernibility relation with a dominance relation. Suppose that to each attribute $q\in Q$ there is associated a preference relation $\succeq_q$, such that, without loss of generality, for all $x,y \in U$, $x$ is at least as good as $y$ with respect to attribute $q$, denoted by $x \succeq_q y$, if $f(x,q) \geq f(y,q)$. Given $x,y \in U$ and $P \subseteq Q$, $x$ \textit{dominates} $y$ with respect to  $P$,
denoted by $x {D_P} y$,  if $f(x,q) \ge f(y,q)$ for all $q \in P$. The  $P$-dominance $D_P$ is a reflexive and
transitive binary relation, i.e., it is a preorder.



%

Given a set of criteria $ P \subseteq Q, P\neq \emptyset$, and $x \in U$, the granules
of knowledge used for approximation in DRSA are the $P$-\textit{dominating set} $D_P^+ \left( x \right)$ and the $P$-\textit{dominated set} $D_P^ - \left( x \right)$ defined as follows:
\[ 
D_P^+ \left( x \right)=\{y \in U: y D_P x\}, \quad D_P^ - \left( x \right)=\{y\in U: x {D_P} y\}.
\]

Given $P \subseteq Q, P\neq \emptyset$, and $X \subseteq U$,  the $P$-upward-lower and the $P$-upward-upper approximation of $X$  are defined  as follows:

$$
\underline{P}^+(X)=\left\{x \in U: D^+_P(x) \subseteq X \right\},
$$
$$
\overline{P}^{\; +}(X)=\left\{x \in U: D^-_P(x) \cap X \neq \emptyset \right\}.
$$
Analogously, the $P$-downward-lower and the $P$-downward-upper approximation of $X$  are defined  as follows:
$$
\underline{P}^-(X)=\left\{x \in U: D^-_P(x) \subseteq X \right\},
$$
$$
\overline{P}^{\; -}(X)=\left\{x \in U: D^+_P(x) \cap X \neq \emptyset \right\}.
$$

Also within DRSA one can consider a division of the set of attributes  $Q$  into the set of condition attributes 
$C\neq\emptyset$ and the set of decision attributes 
$D\neq\emptyset$, so that, taken $P \subseteq C$ and $S \subseteq D, P\neq \emptyset$ and $S\neq \emptyset$, and $x \in U$, the $P$-lower and the $P$-upper approximation of the $S$-dominating set $D^+_S(x)$ and the $S$-dominated set $D^-_S(x)$ can be computed, with the aim of discovering dependencies between condition attributes from $P$ and decision attributes from $S$. Indeed, for example, if for $x,z \in U$, $P\subseteq C$ and $S \subseteq D$, $x \in \underline{P}^+(D^+_S(z))$, then $D^+_P(x) \subseteq D^+_S(z)$, that is, for all $y \in U$, if $f(y,q)\ge f(x,q)$ for all $q \in P$, then $f(y,q) \ge f(z,q)$ for all $q \in S$. This can be interpreted in the sense that the objects from $U$ suggest the following \textit{certain} decision rule

\begin{center}
$\mathbf{\rho}^+_c$=``if $f(y,q)\ge f(x,q)$ for all $q \in P$, then $f(y,q)\ge f(z,q)$ for all $q \in S$''.
\end{center}

Analogously, if for $x,z \in U$, $P\subseteq C$ and $S \subseteq D$, $P\neq \emptyset$ and $S\neq \emptyset$, $y \in \overline{P}^+(D^+_S(z))$, then $D^-_P(y) \cap D^+_S(z) \neq \emptyset$, that is, there is at least one $x \in U$, such that $f(x,q) \le f(y,q)$ for all $q \in P$ and $f(x,q) \ge f(z,q)$ for all $q \in S$. This can be interpreted in the sense that the objects from $U$ suggest the following \textit{possible} decision rule

\begin{center}
$\mathbf{\rho}^{+}_p$=``if $f(y,q)\ge f(x,q)$ for all $q \in P$, then it is possible that $f(y,q) \ge f(z,q)$ for all $q \in S$''.  
\end{center}  

In this context, one has to consider a semantic correlation \cite{greco2002mining} between condition and decision attributes for which an improvement on a condition attribute should not worsen a decision attribute, if the values of the other condition attributes remain unchanged. For example, semantic correlation implies that if the evaluation of a student on a given subject, let us say history, improves, the overall evaluation should not decrease if for all other subjects the evaluations remain the same. Here, history is a condition attribute and the overall evaluation is a decision attribute. Considering semantic correlation, taken $S \subseteq D, S \neq \emptyset$, dominating sets $D^+_S(x)$ admit upward approximations and dominated sets $D^-_S(x)$ admit downward approximations, and, consequently, the following rough approximations can be considered:
$$
\underline{P}^+(D^+_S(x))=\left\{y \in U: D^+_P(y) \subseteq D^+_S(x) \right\},
$$
$$
  \overline{P}^{\; +}(D^+_S(x))=\left\{y \in U: D^-_P(y) \cap D^+_S(x) \neq \emptyset \right\},
$$
$$
\underline{P}^-(D^-_S(x))=\left\{y \in U: D^-_P(y) \subseteq D^-_S(x) \right\}, 
$$
$$
\overline{P}^{\; -}(D^-_S(x))=\left\{y \in U: D^+_P(y) \cap D^-_S(x) \neq \emptyset \right\}.
$$

As was the case for IRSA, also for DRSA, usually, $D=\{d\}$, so that a single decision attribute $d$ is considered. In this case, the decision classes $\{Cl_1,\ldots,Cl_n\}$ can be preferentially ordered so that for all $x,y \in U$, with $x \in Cl_{t_1}$ and $y \in Cl_{t_2}$, if $t_1 \ge t_2$, then  $x \succeq_d y$ and, equivalently, $f(x,d)\ge f(y,d)$. Consequently, the dominating and the dominated sets of the decision attribute $d$, $D^+_d(x)$ and $D^-_d(x)$, $x\in U$, can be formulated in terms of upward and downward unions of decision classes $Cl_t^\ge$ and $Cl_t^\le$, $t=1,\ldots,n$, defined as
\[
Cl_t^\ge=\bigcup_{s\ge t}Cl_s, \quad Cl_t^\le=\bigcup_{s\le t}Cl_s.
\]
In fact, for all $x\in U$, if $x \in Cl_t$, then
\[
D^+_d(x)=\{y\in U: f(y,d) \ge f(x,d)\}=Cl_t^\ge, 
\]
\[
D^-_d(x)=\{y\in U: f(y,d) \le f(x,d)\}=Cl_t^\le.
\]

In this context, taking into consideration semantic correlation, the upward lower and upper approximation $\underline{P}^+(Cl^\ge_t)$ and $\overline{P}^+(Cl^\ge_t)$ of each upward union of decision classes $Cl_t^\ge$ with respect to condition attributes $P \subseteq C, P \neq \emptyset$, can be obtained. Analogously, the downward lower and upper approximation $\underline{P}^-(Cl^\le_t)$ and $\overline{P}^-(Cl^\le_t)$ of each downward union of decision classes $Cl_t^\le$ with respect to condition attributes $P \subseteq C, P \neq \emptyset$, can be computed.


\section{Developments of DRSA}\label{sec:developments}
In this section we discuss some of the many developments and extensions of DRSA, related both to application of DRSA to other types of problems than just ordinal classification, considered in the basic version of DRSA presented in the previous section, and to adaptation of DRSA to different characteristics of analyzed data.

\subsection{DRSA to multicriteria choice and ranking}\label{sec:ranking}
Choice and ranking problems are one of the main problems considered in MCDA \cite{roy1985methodologie}. To address them using rough sets, Greco, Matarazzo and S{\l}owi\'nski proposed the concept of a pairwise comparison table (PCT) \cite{gre_mat_slo_1995,greco1997rough} in which binary relations, that is sets of pairs of objects, are approximated, rather than sets of single objects as in the basic IRSA and DRSA. Although this way it became possible to take into account the preference over particular criteria and the overall preference between two objects, still PCT was analyzed using IRSA, which could not take into account all types of inconsistency observed in pairwise comparisons. This deficiency was overcome when Greco, Matarazzo and S{\l}owi\'nski proposed to employ dominance relation while processing PCT \cite{Gre_Mat_Slo_1996, GMS_1996}.

Over the years, different ways of application of DRSA to multicriteria choice and ranking have been proposed in the literature. In the following, we discuss five such approaches in order of their appearance, denoting them by Greek letters $\alpha-\epsilon$. All these approaches involve five key steps:
\begin{enumerate}
	\item[($s_1$)] elicitation of preference information in terms of pairwise comparisons of some reference objects,
	
	\item[($s_2$)] rough approximation of comprehensive relations implied by the pairwise comparisons, using the DRSA or a Variable Consistency DRSA (VC-DRSA, see Section \ref{sec:VC-DRSA}), to handle possible inconsistencies observed in the PCT,
	
	\item[($s_3$)] induction of decision rules from rough approximations of considered comprehensive relations,
	
	\item[($s_4$)] application of induced decision rules on set $M \subseteq U$ of objects to be ranked,
	
	\item[($s_5$)] exploitation of the resulting preference structure on $M$ to get a final ranking of objects (total preorder).
\end{enumerate}
When dealing with multicriteria ranking problem, this final ranking is the recommendation presented to the DM. In case of multicriteria choice, the recommendation is the object or the set of objects ranked as the best.

A common assumption of approaches $\alpha-\delta$ is that for each cardinal criterion \mbox{$q_i \in C$} (i.e., criterion with a cardinal scale, for which it is meaningful to consider intensity of preference) there is given a set of \emph{graded preference relations} $T_i=\{P_i^h, h \in H_i\}$, where $H_i$ is a finite set of integer numbers (``grades of intensity of preference'') (see, e.g., \cite{greco2001rough}). Relations $P_i^h$ are binary relations over $U$, such that for $a,b \in U$:
\begin{itemize}
	\item if $a P_i^h b$ and $h>0$, then object $a$ is preferred to object $b$ by degree $h$ w.r.t. criterion $q_i$,
	\item if $a P_i^h b$ and $h<0$, then object $a$ is not preferred to object $b$ by degree $h$ w.r.t. criterion $q_i$,
	\item if $a P_i^h b$ and $h=0$, then object $a$ is similar (asymmetrically indifferent) to object $b$ w.r.t. criterion $q_i$.
\end{itemize}

In approach $\epsilon$, however, it is argued that the above modeling of binary relations $P_i^h$, involving determination of several thresholds for each cardinal criterion, may be considered impractical. In fact, it relates to discretization of the scale on which the strength of preference is expressed. Therefore, in approach $\epsilon$, instead of grades of preference intensity, one puts in a PCT for each cardinal criterion $q_i \in C$ just differences of evaluations $f(a,q_i)-f(b,q_i)$.

The first application of DRSA to multicriteria choice and ranking, denoted by~$\alpha$, was proposed in \cite{Gre_Mat_Slo_1996,GMS_1996,Gre_Mat_Slo_1998_PCT,Gre_Mat_Slo_Tso_1998,Greco_et_al_1999} and reminded in \cite{greco1999use,Slowinski_et_al_2009}. It is characterized by the following steps:
\begin{itemize}
	\item[(s$_1^{\alpha}$) ] the pairwise comparisons of reference objects are expressed in terms of outranking and non-outranking relations; given a pair of objects $(a,b) \in U \times U$, a~DM may: (i) state that object~$a$ is comprehensively at least as good as object~$b$ (or, in other words, $a$ outranks $b$), denoted by $a S b$, (ii) state that object~$a$ is comprehensively not at least as good as object $b$ (or, in other words, $a$ does not outrank $b$), denoted by $a S^c b$, or (iii) abstain from any judgment;
	
	\item[(s$_2^{\alpha}$)] relations $S$ and $S^c$ are approximated using \emph{graded dominance relations} (called in the following \emph{single-graded dominance relations}) w.r.t. the set of criteria $C$;
	
	\item[(s$_3^{\alpha}$)] the approximations of $S$ and $S^c$ are used to generate four types of single-graded decision rules (i.e., concerning the same grade of preference w.r.t. each criterion present in the rule condition part), denoted by $D_{++}$, $D_{-+}$, $D_{+-}$, $D_{--}$; if a pair $(a,b) \in U \times U$ is covered by a rule of the first two types, it is concluded that $a S b$, while if it is covered by a rule of the last two types, the conclusion is $a S^c b$;
	
	\item[(s$_4^{\alpha}$)] the application of induced rules on set $M \subseteq U$ yields four outranking relations called \emph{true outranking relation}, \emph{false outranking relation}, \emph{contradictory outranking relation}, and \emph{unknown outranking relation}, which together constitute so-called \emph{four-valued outranking} \cite{Tso_Vin_1995,Tso_Vin_1997};
	
	\item[(s$_5^{\alpha}$)] the final ranking of objects from set $M \subseteq U$ is obtained using their so-called \emph{net flow scores}; the net flow score of an object $a \in M$, denoted by $S^{NF}(a)$, is calculated as the sum of:
	\begin{itemize}
		\item[(i)] the number of objects $b \in M$ such that the induced rules suggest $a S b$, and
		\item[(ii)] the number of objects $b \in M$ such that the induced rules suggest $b S^c a$,
	\end{itemize}
	diminished by the sum of:
	\begin{itemize}
		\item[(iii)] the number of objects $b \in M$ such that the induced rules suggest $b S a$, and
		\item[(iv)] the number of objects $b \in M$ such that the induced rules suggest $a S^c b$.
	\end{itemize}
\end{itemize}
	
It is worth noting that the first approach presented in \cite{Gre_Mat_Slo_1996,GMS_1996,Gre_Mat_Slo_1998_PCT,Gre_Mat_Slo_Tso_1998,Greco_et_al_1999} does not account for ordinal criteria (i.e., criteria with ordinal scale, for which consideration of intensity of preference is not meaningful). Moreover, the single-graded dominance relation is lacking in precision \cite{greco1999use} as it assumes a common grade of intensity of preference for all considered (cardinal) criteria.

The second application of DRSA to multicriteria choice and ranking, denoted by~$\beta$, was presented in \cite{greco1999use,Gre_Mat_Slo_2000_INFOR,greco2001rough,Slowinski_2002,Slowinski_et_al_2005,Slowinski_et_al_2009,Slowinski_et_al_2014,slowinski2015rough}. It comprises of the following steps:
\begin{itemize}
	\item[(s$_1^{\beta}$)] $\equiv$ (s$_1^{\alpha}$);
	
	\item[(s$_2^{\beta}$)] relations $S$ and $S^c$ are approximated using the dominance relation that accounts for both cardinal and ordinal criteria; w.r.t. cardinal criteria, the \emph{multigraded dominance relation} is considered;
	
	\item[(s$_3^{\beta}$)] the approximations of $S$ and $S^c$ are used to generate three types of decision rules (that can use different grades of preference w.r.t. each cardinal criterion present in the rule condition part), denoted by $D_{\ge}$, $D_{\le}$, and $D_{\ge\le}$; if a pair of objects $(a,b) \in U \times U$ is covered by a rule of the first type, it is concluded that $a S b$, while if it is covered by a rule of the second type, the conclusion is $a S^c b$;
	
	\item[(s$_4^{\beta}$)] $\equiv$ (s$_4^{\alpha}$);
	
	\item[(s$_5^{\beta}$)] $\equiv$ (s$_5^{\alpha}$).
\end{itemize}

It is worth noting that definitions of lower approximations applied in approaches $\alpha$ and $\beta$ appear to be too restrictive in practical applications. In consequence, lower approximations of $S$ and $S^c$ are often small or even empty, preventing a good generalization of pairwise comparisons in terms of decision rules.

The third application of DRSA to multicriteria choice and ranking presented in \cite{Slowinski_2002,gre_mat_slo_2005a}, denoted by $\gamma$, is characterized by the following steps:
\begin{itemize}
	\item[(s$_1^{\gamma}$)] $\equiv$ (s$_1^{\beta}$);
	
	\item[(s$_2^{\gamma}$)] relations $S$ and $S^c$ are approximated using the dominance relation that accounts for both cardinal and ordinal criteria; w.r.t. cardinal criteria, the multigraded dominance relation is considered; contrary to step (s$_2^{\beta}$), the approximations of $S$ and $S^c$ are calculated using a PCT-oriented adaptation of the VC-DRSA proposed originally in \cite{gre_mat_slo_ste_2000a} w.r.t. the multicriteria classification problems; as this VC-DRSA measures consistency of decision examples using rough membership measure $\mu$, it will be denoted by $\mu$-VC-DRSA;
	
	\item[(s$_3^{\gamma}$)] the lower approximations of $S$ and $S^c$ are used to generate two types of \emph{probabilistic decision rules} (that can use different grades of preference w.r.t. each cardinal criterion present in the rule condition part), denoted by $D_{\ge}$ and $D_{\le}$; if a pair of objects $(a,b) \in U \times U$ is covered by a rule of the first type, it is concluded that $a S b$, while if it is covered by a rule of the second type, the conclusion is $a S^c b$;
	
	\item[(s$_4^{\gamma}$)] $\equiv$ (s$_4^{\beta}$);
	
	\item[(s$_5^{\gamma}$)] $\equiv$ (s$_5^{\beta}$).
\end{itemize}

The fourth application of DRSA to multicriteria choice and ranking, denoted by~$\delta$, was introduced in \cite{For_Gre_Slo_2008}. It is characterized by the following steps:
\begin{itemize}
	\item[(s$_1^{\delta}$)] the pairwise comparisons of reference objects are expressed in terms of \emph{comprehensive graded preference relations} $\succ^h$, $h \in [-1,1]$; given a pair of objects $(a,b) \in U \times U$, a DM may: (i) state that object $a$ is comprehensively preferred to object $b$ in grade $h$, i.e., $a \succ^h b$ with $h>0$, (ii) state that object $a$ is comprehensively \emph{not} preferred to object $b$ in grade $h$, i.e., $a \succ^h b$ with $h<0$, (iii)~state that object $a$ is comprehensively indifferent to object $b$, i.e., $a \succ^0 b$, or (iv) abstain from any judgment;
	
	\item[(s$_2^{\delta}$)] \emph{upward cumulated preference relations} (upward unions of comprehensive graded preference relations) $\succ^{\ge h}$ and \emph{downward cumulated preference relations} (downward unions of comprehensive graded preference relations) $\succ^{\le h}$ are approximated using the dominance relation that accounts for both cardinal and ordinal criteria; w.r.t. cardinal criteria, the multigraded dominance relation is considered; analogously to step (s$_2^{\gamma}$), the approximations of $\succ^{\ge h}$ and $\succ^{\le h}$ are calculated using a PCT-oriented adaptation of $\mu$-VC-DRSA proposed in \cite{gre_mat_slo_ste_2000a};
	
	\item[(s$_3^{\delta}$)] the lower approximations of $\succ^{\ge h}$ and $\succ^{\le h}$ are used to generate two types of \emph{probabilistic decision rules} (that can use different grades of preference w.r.t. each cardinal criterion present in the rule condition part), denoted by $D_{\ge}$ and $D_{\le}$; each induced rule is additionally characterized by the attained \emph{confidence level}; if a pair of objects $(a,b) \in U \times U$ is covered by a rule of the first type, it~is concluded that $a \succ^{\ge h} b$, while if it is covered by a rule of the second type, the conclusion is $a \succ^{\le h} b$;
	
	\item[(s$_4^{\delta}$)] the application of induced rules on set $M \subseteq U$ yields a graded fuzzy preference relation (of level 2) over $M$; this relation is graded because of different grades of preference, but it is also fuzzy because of different confidence levels of rules matching pairs of objects from $M \times M$;
	
	\item[(s$_5^{\delta}$)] the final ranking of objects from set $M \subseteq U$ is obtained by exploitation of the preference structure on $M$ using either the Weighted Fuzzy Net Flow Score (WFNFS) procedure or a Lexicographic-fuzzy Net Flow Score procedure.
\end{itemize}

It is worth noting that elicitation of preferences in terms of comprehensive graded preference relations $\succ^h$ requires, in general, a greater cognitive effort of a DM. Moreover, it makes exploitation of the preference structure resulting from application of induced decision rules more complex.

It is also important to note that the application of variable consistency model of DRSA considered in approaches $\gamma$ and $\delta$, relying on rough membership consistency measure $\mu$, leads to the situation when calculated lower approximations of considered comprehensive relations lack several desirable monotonicity properties \cite{Bla_Gre_Slo_Sze_2009}.

The fifth application of DRSA to multicriteria choice and ranking, denoted by~$\epsilon$, was introduced in \cite{Sze_Gre_Slo_2013,Sze_Gre_Slo_2014_INS} and extended in \cite{SzelagPhD2015}. In this approach, one takes into account the nature of the set of criteria $C$ and acts accordingly, both during composition of a PCT and when constructing the preference structure resulting from application of decision rules. In \cite{Sze_Gre_Slo_2013}, a typical MCDA assumption is adopted that set $C$ is a \emph{consistent family of criteria} \cite{rb1993}, i.e., it satisfies the properties of \emph{completeness} (all relevant criteria are considered), \emph{monotonicity} (the better the evaluation of an object on considered criteria, the more it is preferable to another object), and \emph{non-redundancy} (there is no criterion which could be removed without violating one of the previous two properties). Let us call this variant by~$\epsilon_c$. In \cite{Sze_Gre_Slo_2014_INS}, the above assumption is dropped, which is typical for Preference Learning \cite{Fur_Hull_2010} methods. Let us call this variant by~$\epsilon_{nc}$ (for not necessarily consistent set of~criteria).

Approach $\epsilon$ consists of the following steps:

\begin{itemize}
	\item[(s$_1^{\epsilon}$)] the preference information is elicited as in step (s$_1^{\gamma}$); 
	moreover, relation $S$ is enriched with some additional pairs of objects -- pairs $(x,y)$ such that $x$ dominates~$y$ in variant $\epsilon_c$, and pairs $(x,x)$ in variant $\epsilon_{nc}$;
	
	\item[(s$_2^{\epsilon}$)] one proceeds as in step (s$_2^{\gamma}$), but a PCT-oriented adaptation of the VC-DRSA proposed in \cite{Bla_Gre_Slo_Sze_2007,Bla_Gre_Slo_Sze_2009} is used instead, employing consistency measure $\epsilon$ (denoted by $\epsilon$-VC-DRSA);
	
	\item[(s$_3^{\epsilon}$)] one proceeds as in step (s$_3^{\gamma}$), but inducing probabilistic decision rules using VC-DomLEM minimal cover algorithm \cite{Bla_Slo_Sze_2011}, employing rule consistency measure $\widehat\epsilon_T$, where $T \in \{S,S^c\}$ \cite{SzelagPhD2015};
	
	\item[(s$_4^{\epsilon}$)] application of induced rules, both suggesting assignment to relation $S$ and $S^c$, on set $M \subseteq U$; this way one gets a preference structure on $M$, composed of relations $\mathbb{S}$ and $\mathbb{S}^c$ over $M$; both relations can be either crisp (when it is only checked if there exists at least one rule suggesting assignment of pair $(a, b) \in M \times M$ to relation $S$/$S^c$) or valued (when it is checked what is the \emph{strength of the strongest rule} suggesting assignment of $(a, b)$ to $S$/$S^c$), and they are enriched with additional pairs of objects depending on the variant $\epsilon_c$ or $\epsilon_{nc}$ \cite{SzelagPhD2015}; moreover, when constructing valued relations, rule strength can depend on rule's consistency only \cite{SzelagPhD2015} or also on rule's coverage factor \cite{Sze_Gre_Slo_2014_INS,SzelagPhD2015};
	
	\item[(s$_5^{\epsilon}$)] crisp/valued preference structure composed of crisp/valued relations  $\mathbb{S}$ and $\mathbb{S}^c$ over $M \subseteq U$ is exploited by transforming it to valued relation $\mathcal{R}$ defined as:
	
	\begin{equation}
		\mathcal{R}(a,b)=\frac{\mathbb{S}(a,b) + (1 - \mathbb{S}^c(a,b))}{2},\label{relationR}
	\end{equation}

	and by applying the well-known Net Flow Rule (NFR) \cite{Bouyssou_1992c, Bouyssou_Vincke_1997} ranking method, possessing desirable properties \cite{SzelagPhD2015}, to produce the final ranking of objects from set $M$ (a weak order).
\end{itemize}

The efficiency of approach $\epsilon$ was proved experimentally in \cite{SzelagPhD2015}. Moreover, it is implemented in two computer programs: ruleRank and RUDE (see Section \ref{sec:ruleRank}).

\subsection{Case-based reasoning using dominance-based decision rules}
Around the year 2005, Greco, Matarazzo and S{\l}owi\'nski noticed potential application od DRSA to Case-based reasoning (CBR) (see, e.g., \cite{Kolodner_1993}). In CBR, also called similarity-based reasoning, the more similar are the causes, the more similar one expects the effects. This monotonic relationship can be employed by DRSA, when applied to similarity-based classification defined as follows. There is given a finite set of objects $U$ (called \emph{universe of discourse}, or \emph{case base}) and a finite family of pre-defined decision classes $\mathbf{Cl}$. An object $y \in U$ (a ``case'') is described in terms of features $f_1, \ldots, f_m \in F$. For each feature $f_i \in F$, there is given a \emph{marginal similarity function} $\sigma_{f_i} : U \times U \to [0,1]$, such that the value $\sigma_{f_i}(y,x)$ expresses the similarity of object $y \in U$ to object $x \in U$ w.r.t. feature $f_i$; the minimal requirement that function $\sigma_{f_i}$ must satisfy is the following: for all $x,y \in U$, \linebreak
$\sigma_{f_i}(y,x)=1$ if and only if $y \textrm{ and }x\textrm{ have the same value of feature }f_i$. Moreover, for each object  $y \in U$ there is given an information concerning (normalized) \emph{credibility} of its membership to each of the considered classes. To admit a graded credibility, each decision class $Cl_t \in \mathbf{Cl}$, $t \in \{1, \ldots, n\}$, is modeled as a fuzzy set in~$U$ \citep{Zadeh_1965}, characterized by membership function $\mu_{Cl_t} : U \to [0,1]$. Thus, each object $y \in U$ can belong to different decision classes with different degrees of membership.
The aim of decision aiding is to present to~a~DM a~recommendation concerning a new object~$z$, in terms of a degree of membership of this object to particular classes.

In similarity-based classification, the key issue is the aggregation of marginal similarities of objects into their comprehensive similarity. Typically, this aggregation is performed using some real-valued aggregation function (involving operators, like weighted $L_p$ norm, min, etc.) (see, e.g., \citep{Dubois_Prade_1998}) which is always arbitrary to some extent. This motivated Greco, Matarazzo and S{\l}owi\'nski to propose an approach that measures comprehensive similarity in a (more) meaningful way, avoiding the use of an aggregation function. An approach of this type, employing an adaptation of DRSA, was proposed for the first time in \citep{Gre_Mat_Slo_2006}, and improved in \citep{Gre_Mat_Slo_2008a,Gre_Mat_Slo_2008b}. In the proposed approach, comprehensive similarity is represented by decision rules concisely characterizing classification examples. These rules are based on the general monotonic relationship $(mr_1)$: ``the more similar is object $y$ to object $x$ w.r.t. the considered features, the greater the membership of $y$ to a given decision class $Cl_t$'', where $Cl_t \in \mathbf{Cl}$. This enabled to obtain a meaningful similarity measure, which is, moreover, resistant to irrelevant (or noisy) features because each decision rule, being a~partial dominance cone in a similarity space, may involve conditions concerning only a~subset of features. As the induced rules employ only ordinal properties of marginal similarity functions, the considered approach is also invariant to ordinally equivalent marginal similarity functions.

Few years later, a new monotonic relationship $(mr_2)$ was formulated in \cite{Sze_Gre_Bla_Slo_2011}: ``the more similar is object $y$ to object $x$ with respect to the considered features, the closer is $y$ to $x$ in terms of the membership to a given decision class $Cl_t$'', where $Cl_t \in \mathbf{Cl}$. As observed in that paper, it is reasonable to consider $(mr_1)$ only if the membership of reference object $x$ to considered class $Cl_t \in \mathbf{Cl}$ takes a maximum value. On the other hand, $(mr_2)$ does not require any assumption about the membership value of reference objects and can be considered as more general. Additionally, the authors of \cite{Sze_Gre_Bla_Slo_2011} extended the previous approach also by proposing the way of induction of decision rules using VC-DomLEM algorithm \cite{Bla_Slo_Sze_2011}, and by indicating a suitable way of application of these rules according to \cite{Bla_Gre_Slo_2007}. Induced rules underline general monotonic relationship between comprehensive closeness of objects and their marginal similarities. An example of obtained decision rule is the following: ``\emph{if}~similarity of flower $y$ to flower $x$ w.r.t. petal length is at least $0.7$, \emph{and} similarity of flower $y$ to flower $x$ with respect to sepal width is at least $0.8$, \emph{then} the membership of $y$ to class $setosa$ is between $0.7$ and $0.9$''.

The approach started in 2011 was further extended in \cite{SzelagPhD2015}, by further formalizing the adaptation of DRSA, and by revising the rule classification scheme described in \cite{Bla_Gre_Slo_2007}. Finally, the complete approach was presented at a rough set conference in \cite{Sze_Gre_Slo_2016}, where the description of revised rule classification scheme was much simplified.

\subsection{Adaptations of DRSA to handle missing attribute values}\label{sec:missingValues} 

Shortly after the introduction of DRSA, Greco, Matarazzo, and S{\l}owi\'nski considered two extensions of DRSA that enabled analysis of classification data with missing attribute values \cite{Gre_Mat_Slo_1999b,gre_mat_slo_2000}. Also other authors considered this problem \cite{Hu_Liu_2007,Yang_et_al_INS_2008,Dembczynski_et_al_2005,Dembczynski_et_al_2009,Liang_at_al_2010,Bla_Slo_Sze_2012,Du_Hu_2016_INS,Sze_Bla_Slo_2017}. This research, apart from \cite{Du_Hu_2016_INS}, is well summarized in \cite{Sze_Bla_Slo_2017}, where all the approaches are given an id, and their properties are thoroughly examined with respect to a list of 11 desirable properties.

An adaptation of DRSA to handle missing attribute values involves an adjusted definition of dominance relation, that accounts for missing values. In some approaches, it also involves the change of the definition of rough approximation, to take into account the lack of some classical properties of redefined dominance relation (e.g., lack of transitivity). In \cite{Sze_Bla_Slo_2017}, the authors denoted each adaptation identified in the literature by DRSA-$mv_j$, and respective adjusted dominance relation by $D_j$, where $j$ stands for the version id. In the following, we will use $j$ in the superscript, to allow a subset of attributes $P \subseteq C$ in the subscript. The authors of \cite{Sze_Bla_Slo_2017} pointed out, after \cite{Yang_et_al_INS_2008,Slo_Van_2000}, that the following generalized definitions of $P$-lower and $P$-upper approximations of unions of decision classes $Cl_t^\ge, Cl_t^\le$, $t=1,\ldots n$, should be employed:

\begin{equation}\label{eq:generalizedApproximations}
	\arraycolsep=2pt\def\arraystretch{2.0}
	\begin{array}{ll}
		\underline{P}(Cl_t^\ge)=\{x \in U: {d_P^{j+}(x)} \subseteq Cl_t^\ge\} & \ \ \underline{P}(Cl_t^\le)=\{x \in U: D_P^{j-}(x) \subseteq Cl_t^\le\}\\
		\overline{P}(Cl_t^\ge)=\bigcup\limits_{x \in Cl_t^\ge} D_P^{j+}(x) & \ \ \overline{P}(Cl_t^\le)=\bigcup\limits_{x \in Cl_t^\le} {d_P^{j-}(x)},
	\end{array}
\end{equation}
where
\begin{equation}\label{eq:d_j}
	{d_P^{j+}(x)}=\{y \in U : x\,d_P^j\,y\} \ \ \ \ \ \ \ {d_P^{j-}(x)}=\{z \in U : z\,d_P^j\,x\}.
\end{equation}
In the above definitions, $D_P^j$ denotes adapted $P$-dominance relation ($x D_P^j y$ means $x$ $P$-dominates $y$), while $d_P^j$ denotes adapted $P$-inverse-dominance relation ($x d_P^j y$ means $x$ is $P$-dominated by $y$). These dominance relations are defined as:
\begin{align}
	y D_P^j x \Leftrightarrow {y \succeq_{q}^j x} \textrm{ for each } q \in P\\
	z\,d_P^j\,x \Leftrightarrow {z \preceq_{q}^j x} \textrm{ for each } q \in P\label{eq:inverse-dominance},
\end{align}
where $x,y,z \in U$ and versions of DRSA-$mv_j$ differ by definitions of relations $\succeq_{q}^j$ and~$\preceq_{q}^j$. Remark that definition \eqref{eq:generalizedApproximations} can be applied even when relations $D_P^j$ and $d_P^j$ are only assumed to be reflexive. In particular, it fits the case when $y D_P^j x$ is not equivalent to $x d_P^j y$.

In \cite{Sze_Bla_Slo_2017}, the analysis of properties of different adaptations of DRSA was presented. It involves the following adaptations: \cite{Gre_Mat_Slo_1999b,gre_mat_slo_2000} -- introduces DRSA-$mv_1$ and DRSA-$mv_2$, \cite{Hu_Liu_2007} -- presents DRSA-$mv_{2.5}$, \cite{Yang_et_al_INS_2008} -- proposes DRSA-$mv_{1.5}$, \cite{Dembczynski_et_al_2005,Dembczynski_et_al_2009} -- introduces the concept of a lower-end dominance relation used in DRSA-$mv_4$, and the concept of an upper-end dominance relation resulting in DRSA-$mv_5$, \cite{Liang_at_al_2010} -- presents DRSA-$mv_6$, and \cite{Bla_Slo_Sze_2012} -- defines DRSA-$mv_3$. The analysis resulted in a conclusion that the only non-dominated approaches are DRSA-$mv_{1.5}$, DRSA-$mv_2$, DRSA-$mv_4$, and DRSA-$mv_5$.

It is important to note, that taking into account the semantics of missing values considered, e.g., in \cite{Grzymala_Hu_2001,Grzymala_2011} (and in \cite{Ste_Tso_2001}), it can be said that DRSA-$mv_{1.5}$ treats missing values as ``lost'' (``absent'') values, while DRSA-$mv_2$ treats missing values as ``do not care'' values. These approaches are defined as:
\begin{align}
	{y \succeq_{q}^{1.5} x} \Leftrightarrow y \succeq_{q} x, \textrm{ or } f(y,q) = ? \ \ \ \ \ \ \ \ \ & {z \preceq_{q}^{1.5} x} \Leftrightarrow x \succeq_{q} z, \textrm{ or } f(z,q) = ?\\
	{y \succeq_{q}^2 x} \Leftrightarrow y \succeq_{q} x, \textrm{ or } f(y,q) = *, \textrm{ or } f(x,q) = * \ \ \ \ \ \ \ \ \ & {z \preceq_{q}^2 x} \Leftrightarrow x \succeq_{q} z, \textrm{ or } f(z,q) = *,\nonumber\\
	& \ \ \ \ \ \ \ \ \ \ \ \ \ \ \ \ \textrm{or } f(x,q) = *,
\end{align}
where $x,y,z \in U$, $q \in C$, $?$ denotes a ``lost'' missing value, and $*$ denotes a ``do not care'' missing value.

In \cite{Du_Hu_2016_INS}, Du and Hu proposed so-called characteristic-based dominance relation with ``do not care'' values and ``lost'' values coexisting. It relates to the characteristic relation considered in IRSA \cite{Grzymala_2004,Grzymala_2005}. Let denote this approach by DRSA-$mv_{1.5\&2}$. Then 
\begin{align}
	{y \succeq_{q}^{1.5\&2} x} \ \ \Leftrightarrow \ \ f(y,q) = ?, \textrm{ or } y \succeq_{q} x, \textrm{ or } f(y,q) = *, \textrm{ or } f(x,q) = *.
\end{align}
Although the idea of considering  ``do not care'' values and ``lost'' values simultaneously is worth considering, the surprising part of \cite{Du_Hu_2016_INS} is that the authors do not employ $P$-inverse-dominance relation $d_P^{1.5\&2}$, $P \subseteq C$, determined as in \eqref{eq:inverse-dominance} with relation $\preceq_{q}^{1.5\&2}$ defined as:
\begin{align}
	{z \preceq_{q}^{1.5\&2} x} \ \ \Leftrightarrow \ \ f(z,q) = ?, \textrm{ or } x \succeq_{q} z, \textrm{ or } f(z,q) = *, \textrm{ or } f(x,q) = *,
\end{align}
where $x,y,z \in U$, $q \in P$. Instead, Du and Hu consider that $x d_P^{1.5\&2} y \Leftrightarrow y D_P^{1.5\&2} x$. In our opinion, this is not a proper realization of the idea outlined by S{\l}owi\'nski and Vanderpooten \cite{Slo_Van_2000} (even reminded in \cite{Du_Hu_2016_INS}) who claimed than for set $X \subseteq U$ and reflexive binary relation $R$ over $U$, the lower approximation of set $X$ with respect to $R$ should be calculated as:
\begin{equation}
	\underline{R}(X) = \{x \in U : R^{-1}(x) \subseteq U\},
\end{equation}
where $R^{-1}(x) = \{y \in U : x R y\}$.

\subsection{Extensions of DRSA for interval evaluations on criteria}

One of the ways of handling imprecision in object's evaluation is to use interval evaluations. Interval evaluations may also occur when replacing missing values with a range of possible evaluations. Yet another scenario concerns the case of hierarchical set of attributes, when a range of decisions obtained at some lower level of attribute hierarchy is translated to an interval of evaluations for an upper level criterion \cite{Dem_Gre_Slo_2002}. Motivated by these observations, Dembczy\'nski, Greco and S{\l}owi\'nski proposed some extensions of the classical DRSA, that permit to analyze data with interval evaluations on criteria \cite{Dem_Gre_Slo_2002,Dem_Gre_Slo_2003,Dembczynski_et_al_2005,Dembczynski_et_al_2009}, and also with interval assignments to decision classes \cite{Dem_Gre_Slo_2003,Dembczynski_et_al_2005,Dembczynski_et_al_2009}. All these extensions involved so-called \emph{$P$-possible dominance relation} $\overline{D}_P$. Assuming, without loss of generality, that each criterion is of gain type, $P$-possible dominance relation is defined as:
\begin{equation}
	x \overline{D}_P y \Leftrightarrow u(x,q) \ge l(y,q) \textrm{ for all } q \in P,
\end{equation}
where $P \subseteq C$, and $u(x,q)$, $l(y,q)$ denote, respectively, upper limit of the interval for object $x$ on attribute $q$, and lower limit of the interval for object $y$ on attribute $q$.

In \cite{Dembczynski_et_al_2005,Dembczynski_et_al_2009}, the authors additionally considered two other $P$-dominance relations, $P \subseteq C$. Assuming, without loss of generality, that each criterion is of gain type, they are defined as follows:
\begin{itemize}
	\item \emph{$P$-lower-end dominance relation} $D_P^l$:
	\begin{equation}
		x D_P^l y \Leftrightarrow l(x,q) \ge l(y,q) \textrm{ for all } q \in P,
	\end{equation}
	\item \emph{$P$-upper-end dominance relation} $D_P^u$:
	\begin{equation}
		x D_P^u y \Leftrightarrow u(x,q) \ge u(y,q) \textrm{ for all } q \in P.
	\end{equation}
\end{itemize}
The above papers also took into account interval assignments to decision classes. Suppose $d \in Q$ is a decision criterion. Then, the following rough approximations of upward and downward unions of decision classes are considered:
\begin{align}
	\underline{Cl}_t^\ge = \{y \in U : l(y,d) \ge t\},\\
	\underline{Cl}_t^\le = \{y \in U : u(y,d) \le t\},\\
	\overline{Cl}_t^\ge = \{y \in U : u(y,d) \ge t\},\\
	\overline{Cl}_t^\le = \{y \in U : l(y,d) \le t\}.
\end{align}
Finally, the above rough approximations are the basis for second-order dominance-based rough approximations \cite{Dembczynski_et_al_2005,Dembczynski_et_al_2009}:
\begin{align}
	\underline{P}(\underline{Cl}_t^\ge)=\{x \in U : \overline{D}_P^+(x) \subseteq \underline{Cl}_t^\ge\},\\
	\underline{P}(\underline{Cl}_t^\le)=\{x \in U : \overline{D}_P^-(x) \subseteq \underline{Cl}_t^\le\},\\
	\overline{P}(\overline{Cl}_t^\ge)=\{x \in U : \overline{D}_P^-(x) \cap \overline{Cl}_t^\ge \ne \emptyset\},\\
	\overline{P}(\overline{Cl}_t^\le)=\{x \in U : \overline{D}_P^+(x) \cap \overline{Cl}_t^\le \ne \emptyset\}.
\end{align}

DRSA with interval evaluations has also been considered in \cite{Yang_et_al_2009}, where $P$-possible dominance relation has been used for condition criteria. In that paper, the authors also consider intervals with one or two missing limits. Assuming that each $q \in C$ is of gain type, a missing lower limit is then replaced with the minimal value in the value set of the respective criterion, while missing upper limit is replaced with the maximal value in the value set of the respective criterion.

\subsection{Extending DRSA to address hierarchical structure of attributes}
In hierarchical classification/sorting problems, where the set of attributes has a hierarchical structure, the main difficulty is the propagation of inconsistencies along the tree structure,  i.e., taking into consideration at each node of the tree the inconsistent information from lower level nodes. In \cite{Dem_Gre_Slo_2002}, the inconsistencies are propagated bottom-up, in the form of subsets of possible attribute values (for hierarchical regular attributes), and in the form of intervals of possible values (for hierarchical criteria). Decision rules are induced at each node of the tree. The classification/sorting of new objects is also done from the bottom to the top of the hierarchy, to make the final decision in the root node of the tree.

The extension of DRSA proposed in \cite{Dem_Gre_Slo_2002} assumes that object $y$ $P$-dominates object $x$, $P \subseteq C$, if:
\begin{itemize}
	\item for each regular attribute $q \in P$, the subset of attribute values for object $y$ has a non-empty intersection with the subset of attribute values for object $x$,
	\item for each criterion $q \in P$, $u(x,q) \ge l(y,q)$, where $u(x,q)$, $l(y,q)$ denote, respectively, upper limit of the interval for object $x$ on attribute $q$, and lower limit of the interval for object $y$ on attribute $q$.
\end{itemize}
 

\subsection{Extensions of DRSA for non-ordinal data}
B{\l}aszczy\'nski, Greco, and S{\l}owi\'nski proposed an approach to induction of laws from data, which makes use of the concept of monotonic relationships between values of condition and decision attributes, without assuming its direction a~priori and allowing local monotonicity relationships in subregions of the evaluation space~\cite{bla_gre_slo_2012}. This approach is able to discover local and global monotonicity relationships existing in data. The relationships are represented by monotonic decision rules. To enable the discovery of monotonic rules, a non-invasive transformation of the input data was proposed. The proposed transformation should be applied to all non-ordinal attributes. Moreover, after transformation of input data, DRSA is applied to structure data into consistent and inconsistent parts.

For the purpose of the illustration we may assume, without loss of generality, that the value sets of both decision attribute (class labels) and condition attributes are number-coded. As in non-ordinal classification problems the natural complete ordering of classes $Cl_1$, $Cl_2$, $\ldots$, $Cl_n$ induced by number-coded class labels is not entering, in general, into some monotonic relationships with value sets of condition attributes, we have to consider $n$ binary ordinal classification problems with two sets of objects: class $Cl_t$ and its complement $\lnot Cl_t$, $t = 1$, $\ldots$, $n$, which are number-coded by 1 and 0, respectively. This means that in the $t$-th ordinal binary classification problem, set $Cl_t$ is interpreted by DRSA as union $Cl_1^{\ge}$ and set $\lnot Cl_t$ as union $Cl_0^{\le}$, $t = 1$, $\ldots$, $n$. Ordinal classification problems  can be handled by DRSA without altering the original number codes of the class labels.

The transformation of each non-ordinal condition attribute from $C$ is made individually, depending on its type: numerical (number-coded) or nominal. Each numerical (number-coded) attribute $q_i$ is represented, in the transformed form, as a pair of ordinal attributes $q_i'$, and $q_i''$. In other words, evaluation of each object $x \in  U$ by numerical attribute $q_i$ is repeated twice, and the first evaluation $f(x,q_i)$ is renamed to $f(x,q_i')$, while the second evaluation $f(x,q_i)$ is renamed to $f(x,q_i'')$. Then, the first attribute $q_i'$ is set to have positive monotonic relationship with (possibly transformed) decision attribute, while the second attribute $q_i''$ is set to have negative monotonic relationship with the decision attribute.

Each nominal attribute $q_j$ with value set composed of $k$ distinct values is binarized, such that the presence or absence of the $l$-th value of this attribute is coded by a new ordinal attribute $q_{jl}$ taking value 1 or 0, respectively, $l = 1$, $2$, $\ldots$, $k$. Then the binary attribute $q_{jl}$ is represented, in the transformed form, by a pair of ordinal attributes $q_{jl}'$, and $q_{jl}''$. Again, the first ordinal attribute in that pair has positive monotonic relationship with (possibly transformed) decision attribute, while the second attribute in that pair has negative monotonic relationship.

The proposed approach provides framework for analysis of heterogeneous classification data. It has been shown experimentally in~\cite{bla_gre_slo_2012}, that the monotonic rules induced from transformed data, together with a specific classification scheme, have at least as good predictive ability as other well known predictors.

\subsection{Parametric, decision theoretic, and stochastic DRSA} \label{sec:VC-DRSA}

Greco, Matarazzo, S{\l}owi\'nski, and Stefanowski identified the need to relax the definition of the lower approximation of union of classes and, in consequence, to admit to the lower approximation some inconsistent objects (i.e., objects, which would not be admitted to lower approximations in classical DRSA) for which there is enough evidence for their membership to the union of classes~\cite{gre_mat_slo_ste_2000a}. In this way, lower approximations are defined assuming an acceptable value of a measure expressing evidence of membership to the set. Following this idea, the evidence for this membership may be estimated by different types of consistency or precision measures, and lower approximations of unions of decision classes may be defined in different ways resulting in different approaches to this kind of relaxation, called generally parametric DRSA, including: Variable Consistency DRSA (VC-DRSA)~\cite{gre_mat_slo_ste_2000a,Bla_Gre_Slo_Sze_2006,Bla_Gre_Slo_Sze_2007,Bla_Gre_Slo_Sze_2009}, \cite{GMS_2001,gre_mat_slo_2005a}, and Variable Precision DRSA (VP-DRSA)~\cite{inuiguchi_2006a,Gre_Mat_Slo_2008b}. For example, in case of VC-DRSA, lower approximation is defined as a subset of the approximated set.
In consequence, given an upward (downward) union of classes, objects that do not belong to this union are never included in its lower approximation, even if they dominate (are dominated by) an object from the considered union.
On the other hand, in case of VP-DRSA, a lower approximation is not a subset of the approximated set. All objects belonging to dominance cones of objects from a lower approximation of a union of classes are also included in the lower approximation of the union.

The fact that precision measure can be interpreted as a conditional probability stimulated development of the decision-theoretic rough set model (DTRSM)~\cite{Yao08,Yao2016}. DTRSM for DRSA has also been studied in~\cite{Greco07,Liu2017}. DTRSM connects definition of approximations with conditional risk minimization in Bayesian decision theory. In that situation, the states correspond to the decision classes, and assignment of an object to positive, negative and boundary regions is decided on the basis of its condition attribute values. Assuming an acceptable loss of the classification accuracy and estimating conditional probability of the considered decision classes, an optimal Bayes decision rule is obtained. In this way, calculation of approximations can be seen as a classification problem.

Another type of approach to treatment of inconsistent objects within DRSA has also been considered in~\cite{kot_slo_2008,kotlowski_2009}. This approach originates from statistical learning and statistical decision theory and, in contrast to the previously mentioned approaches, involves relabeling of objects (i.e., change of class to which object belongs to the more probable one). It uses the notion of stochastic dominance and maximum likelihood estimators of probability of object belonging to union of classes. Stochastic lower approximations of unions of classes are composed of objects for which values of estimators are higher than a given threshold.

It is worth mentioning that statistical interpretations of VC-DRSA and VP-DRSA were proposed in~\cite{ksu_bla_inu_slo_2021}, by connecting lower approximations with minimizers of empirical risk functions. In result, it has been demonstrated that families of monotonic classifiers and the hinge loss function serve as a foundation for characterization of the parametric DRSA with consistency measures having desirable monotonicity properties.


\subsection{Decision rules induction}\label{sec:rules} 

In DRSA, induction of decision rules is subsequent to computation of rough approximations. In this context, computation of approximations can be viewed as a kind of preprocessing of data. Objects identified as sufficiently consistent are a good basis for induction of decision rules. The purpose of induction of decision rules is to discover strong relationships between description of these objects and their membership to a union of classes. If the rules are intended to be used in classification, then the goal of the induction procedure is to find a preferably small set of rules with high predictive accuracy. Decision rules should be short and accurate. Shorter decision rules are easier to understand. Shorter rules also allow to avoid \emph{overfitting} the training data. Overfitting occurs when the learned model fits training data perfectly but is not performing well on new data.

In DRSA, for a given class $Cl_t$, $t \in \{1,\ldots,n\}$, we consider decision rules of the type:
\begin{eqnarray}
\textrm{\emph{if} } f(y,q_{i_1}) \succeq r_{i_1} \land \ldots \land f(y,q_{i_p}) \succeq r_{i_p} \land f(y,q_{i_{p+1}}) = r_{i_{p+1}} \land \ldots \land f(y,q_{i_z}) = r_{i_z} \nonumber \textrm{\ \ \ \ \ }
\\
\textrm{\emph{then} }y \in Cl_t^{\ge} \textrm{,\ \ \ \ \ }
\\
\textrm{\emph{if} } f(y,q_{i_1}) \preceq r_{i_1} \land \ldots \land f(y,q_{i_p}) \preceq r_{i_p} \land f(y,q_{i_{p+1}}) = r_{i_{p+1}} \land \ldots \land f(y,q_{i_z}) = r_{i_z} \nonumber \textrm{\ \ \ \ \ }
\\
\textrm{\emph{then} }y \in Cl_t^{\le} \textrm{,\ \ \ \ \ }
\end{eqnarray}
where $q_j, j \in \{i_1,i_2,\ldots, i_p\}$ denotes a criterion and $q_k, k \in \{i_{p+1},i_{p+2},\ldots,i_z\}$ denotes a regular attribute. Moreover, $r_j \in V_{j}$, $j=\{i_1, i_2, \ldots, i_p, i_{p+1}, i_{p+2}, \ldots, i_z\}$, denotes a value from the domain of criterion / regular attribute $q_j$. We use symbols $\succeq$ and $\preceq$ to indicate weak preference w.r.t. single criterion and inverse weak preference, respectively. If $q_j \in C$ is a gain (cost) criterion, then elementary condition $f(y,q_j) \succeq r_j$ means that the value of covered object $y$ on criterion $q_j$ is not smaller (not greater) than value $r_j$. Elementary conditions for regular attributes are of type $f(y,q_{j}) = r_{j}$.

The most important decision rule induction algorithms in DRSA are inspired by LEM2 algorithm~\cite{grzymala_busse_1992}, proposed by Grzyma{\l}a-Busse. The applied heuristic strategy of rule induction in these algorithms is called sequential covering \citep{han_kam_2006} or separate and conquer \citep{michalski_1969,pag_hau_1990,furnkranz_1999}. It constructs a rule that covers a subset of training objects, removes the covered objects from the training set, and iteratively learns another rule that covers some of the remaining objects, until no uncovered objects remain. The first of such algorithms was proposed by Greco, Matarazzo, S{\l}owi\'nski, and Stefanowski and is called DomLEM~\cite{gre_mat_slo_ste_2000}. VC-DomLEM, proposed by B{\l}aszczy\'nski, Szel\k{a}g, and S{\l}owi\'nski~\citep{Bla_Slo_Sze_2011}, is an adaptation of the same concept for parametric DRSA. This algorithm heuristically searches for rules that satisfy constraint with respect to chosen rule consistency measure. 

\section{Available software}\label{sec:software}
In this section, we describe some computational libraries and applications that implement DRSA and its extensions for rough set analysis of ordinal data.

\subsection{ruleLearn}
ruleLearn\footnote{https://github.com/ruleLearn/rulelearn} is an open-source computation library written in Java and hosted on GitHub (project started in Dec 2017). Currently it implements DRSA and VC-DRSA (see Section \ref{sec:VC-DRSA}), and allows to handle missing attribute values according to DRSA-$mv_{1.5}$ and DRSA-$mv_2$ (see Section \ref{sec:missingValues}). It also offers access to the VC-DomLEM rule induction algorithm (see Section \ref{sec:rules}) and different classification strategies. ruleLearn is utilized by other programs described in this section: RuLeStudio and RuleVisualization.


\subsection{RuLeStudio}
RuLeStudio\footnote{https://github.com/dominieq/rule-studio} is an open-source client-server web application supporting data analysis using DRSA and VC-DRSA. It requires Java 8+ to run, and employs ruleLearn API. This application was finished in 2020 \cite{RuLeStudio_thesis_2020}. It supports application of (VC-)DRSA to analysis of ordinal data, possibly containing missing values (handled according to DRSA-$mv_{1.5}$ or DRSA-$mv_2$). The application features data editor, presentation of dominance cones (both with respect to dominance and inverse-dominance relation), certain/possible rule generation according to VC-DomLEM algorithm, analysis of different characteristics of induced rules, application of rules on test objects using several classification strategies, cross-validation, and presentation of misclassification matrix.

\subsection{RuleVisualization}
RuleVisualization\footnote{http://www.cs.put.poznan.pl/mszelag/Software/RuleVisualization/RuleVisualization.html} is an open-source client-server web application for visualization and exploration of decision rules. It requires Java 8+ to run, and employs ruleLearn API. This application was finished in 2019 \cite{RuleVisualization_thesis_2019}. This program can read rules typical for DRSA, induced, e.g., by RuLeStudio, and offers different visualizations of these rules (including the attributes used by rules), their filtering, sorting, matching to some test objects, and presentation of graphs showing co-occurence of attributes and semantic/coverage similarity among rules.

\subsection{jMAF}
jMAF\footnote{http://www.cs.put.poznan.pl/jblaszczynski/Site/jRS.html} is a rough set data analysis software written in Java. It makes use of java Rough Set (jRS) library. jMAF and jRS library implement methods of data analysis provided by the DRSA and VC-DRSA, allowing to calculate lower approximations, induce decision rules, and apply the rules to classify objects. More details with regard to the analysis that this software package enables can be found in~\cite{bla_et_al_2013}. 

\subsection{RuleRank Ultimate Desktop Edition}\label{sec:ruleRank}
RuleRank Ultimate Desktop Edition (RUDE)\footnote{http://www.cs.put.poznan.pl/mszelag/Software/ruleRank/ruleRank.html} is a decision support tool for application of DRSA to multicriteria choice and ranking. It requires Java 10+ and JavaFX to run. This application was finished in 2018 \cite {RUDE_thesis_2018}. RUDE employs approach denoted in Section \ref{sec:ranking} by $\epsilon$, allowing setting different parameters characteristic for that approach and visualization of: PCT, decision rules induced from that PCT, preference graph resulting from application of these rules on a set of objects to be ranked, and final ranking.



\section{Conclusions}\label{sec:conclusions}

We presented the main ideas and developments of the Dominance-based Rough Approach (DRSA), recalling the basic principles with a general overview of its developments and software.  At the end of this survey, we have to conclude that DRSA continues to be appealing for experts in Multiple Criteria Decision Aiding (MCDA) on one hand, and artificial intelligence and machine learning on the other hand, for its most salient characteristic consisting in its capacity of asking the decision maker for simple preference information and supplying easily understandable and explainable recommendations. It is not difficult to see that the DRSA properties of being simple and easily understandable have to be acknowledged to the contribution of Roman S{\l}owi\'nski following the lines traced by Bernard Roy with respect to MCDA and Zdzis{\l}aw Pawlak with respect to rough set theory.

 \vspace{0.5cm}

\section*{Acknowledgements}
\addcontentsline{toc}{section}{\numberline{}Acknowledgements}
\noindent This research was partially supported by TAILOR, a project funded by EU 
Horizon 2020 research and innovation programme under GA No 952215.
Salvatore Greco wishes to acknowledge the support of the Ministero dell’Istruzione, dell’Universit\'a e della Ricerca (MIUR) - PRIN 2017, project “Multiple Criteria Decision Analysis and Multiple Criteria Decision Theory”, grant
2017CY2NCA and  the research project “Data analytics for entrepreneurial ecosystems, sustainable development and well being indices” of the Department of Economics and Business of the University of Catania.

\bibliographystyle{spbasic}
\bibliography{IntelligentDecisionSupportSystems2022Ch18}

\end{document}